\documentclass[cameraready]{Interspeech}

\title{Mixture of Phonetic Experts Based Low-Rank Adaptation of Conformer Models for Accented English Speech Recognition}

\author[affiliation={1}, orcid={0009-0007-7051-1235}]{Santosh}{Dahal}

\author[affiliation={2}, orcid={0009-0009-8491-8663}]{Anmol}{Guragain}

\author[affiliation={3},orcid={0000-0003-3472-0703}]{Tianchi}{Liu}

\author[affiliation={2}, orcid={0000-0002-3411-7384}]{Luis Fernando}{D'Haro}

\author[affiliation={1}]{Kiran Chandra}{Dahal}

\address{
    $^1$ Department of Electronics and Computer Engineering, Thapathali Campus, Nepal \\
    $^2$ ETSI de Telecomunicaci\'{o}n, IPTC, Universidad Polit\'{e}cnica de Madrid, Spain \\
    $^3$ National University of Singapore, Singapore
}

\email{
\{santosh.080msise16,dahalkc\}@tcioe.edu.np,
\{anmol.g,luisfernando.dharo\}@upm.es,
tianchi\_liu@u.nus.edu
}

\usepackage{comment}

\begin{document}

\maketitle

\begin{abstract}
We present Mixture of Phonetic Experts based Low-Rank Adaptation (MoPE-LoRA), a parameter-efficient adaptation framework for accented English speech recognition. Recent mixture-of-experts approaches typically allocate experts per accent, implicitly treating accent identity as the primary axis of variation. In contrast, we organize adaptation along phonetic categories, motivated by the observation that accent differences are often expressed as systematic distortions of phoneme realization. MoPE-LoRA introduces a fixed set of six low-rank experts aligned with manner-of-articulation classes and performs frame-level routing through a hybrid mechanism that combines phoneme supervision with learned acoustic gating. Because the expert set is shared across accents, the model does not require accent-specific modules and naturally extends to previously unseen accents. Experiments on L2-ARCTIC show that MoPE-LoRA consistently improves over single LoRA and full fine-tuning. In zero-shot evaluation where one accent is held out during training, the proposed method yields a 12.3\% relative improvement over single LoRA. These results indicate that phonetic-category-based decomposition offers a scalable and transferable alternative to accent-level adaptation.
\end{abstract}

\section{Introduction}

While modern ASR systems achieve high accuracy on native English speech~\cite{gulati2020conformerconvolutionaugmentedtransformerspeech}, their performance often degrades on non-native or accented English, particularly in low-resource scenarios~\cite{javed2023svarahevaluatingenglishasr, Zhao2018L2ARCTIC}. Accented speech typically involves systematic shifts in pronunciation and phoneme realization, which models trained primarily on native speech do not fully capture. Improving robustness to such variation remains an important challenge.

A range of adaptation strategies have been explored. Domain adversarial training~\cite{sun2018domainadversarialtrainingaccented, ganin2016domainadversarialtrainingneuralnetworks} aims to learn accent-invariant representations, while multi-task learning~\cite{Accent_Embeddings_Jain_2018} introduces auxiliary objectives such as accent classification or phoneme prediction. While these methods can partially mitigate performance drops, they often require carefully balanced training objectives and additional annotated data, which may not be feasible in low-resource accented speech scenarios.

More recently, parameter-efficient fine-tuning (PEFT) methods, particularly low-rank adaptation (LoRA)~\cite{hu2021loralowrankadaptationlarge, WD_LoRA_Joseph_2024}, have become attractive for adapting large pre-trained ASR models with limited additional parameters. However, standard LoRA applies a single adaptation mechanism across all phonetic contexts, without explicitly modeling how accent variation manifests at the phoneme level~\cite{song2024lorawhisperparameterefficientextensiblemultilingual}.

Mixture-of-Experts (MoE) approaches provide another direction by routing inputs through specialized sub-modules~\cite{Mixture_LORA_Bagat_2025, you2021speechmoe2mixtureofexpertsmodelimproved, hu2023mixtureofexpertconformerstreamingmultilingual, wang2023language}. In multi-accent settings, experts are often assigned per accent or language. While effective when accent identity is known, this formulation treats accent as the primary axis of variation and may require the number of experts to grow with the number of supported accents. In practice, however, accent differences are largely expressed through systematic distortions of phonetic realizations, such as vowel shifts or consonant substitutions, which may recur across multiple accents. Modeling adaptation strictly at the accent level may therefore overlook the shared phonetic structure underlying accent variability.

Motivated by this observation, we propose MoPE-LoRA, a phonetic routing-based adaptation framework for Conformer ASR models. Inspired by Switch Transformer \cite{fedus2022switchtransformersscalingtrillion} and ST-MoE \cite{zoph2022stmoedesigningstabletransferable}, our method introduces phoneme-aware expert routing, allowing the model to dynamically select experts based on phonetic content rather than accent or language identity. By integrating LoRA in selective layers with phonetic routing, our method adapts to accented speech using a fixed set of six low-rank experts aligned with manner-of-articulation classes~\cite{ladefoged2011course}. To our knowledge, this is the first use of phoneme-category-grounded LoRA experts with frame-level hybrid routing for non-native multi-accent ASR.

We evaluate our approach on the L2-ARCTIC dataset~\cite{Zhao2018L2ARCTIC}, a widely used benchmark for non-native English speech recognition. Our method improves WER over single LoRA adaptation, demonstrating that phoneme-aware MoE with phonetic routing provides an accurate, interpretable, and parameter-efficient solution for accented ASR. To further assess generalization, we conducted unseen-accent experiments by holding out one accent from the L2-ARCTIC dataset for testing while training on the remaining accents. Across all evaluated accents, our approach achieved the lowest error rates, highlighting its robustness to previously unseen accent variations.

\begin{figure}[t]
  \centering
  \includegraphics[width=0.77\linewidth]{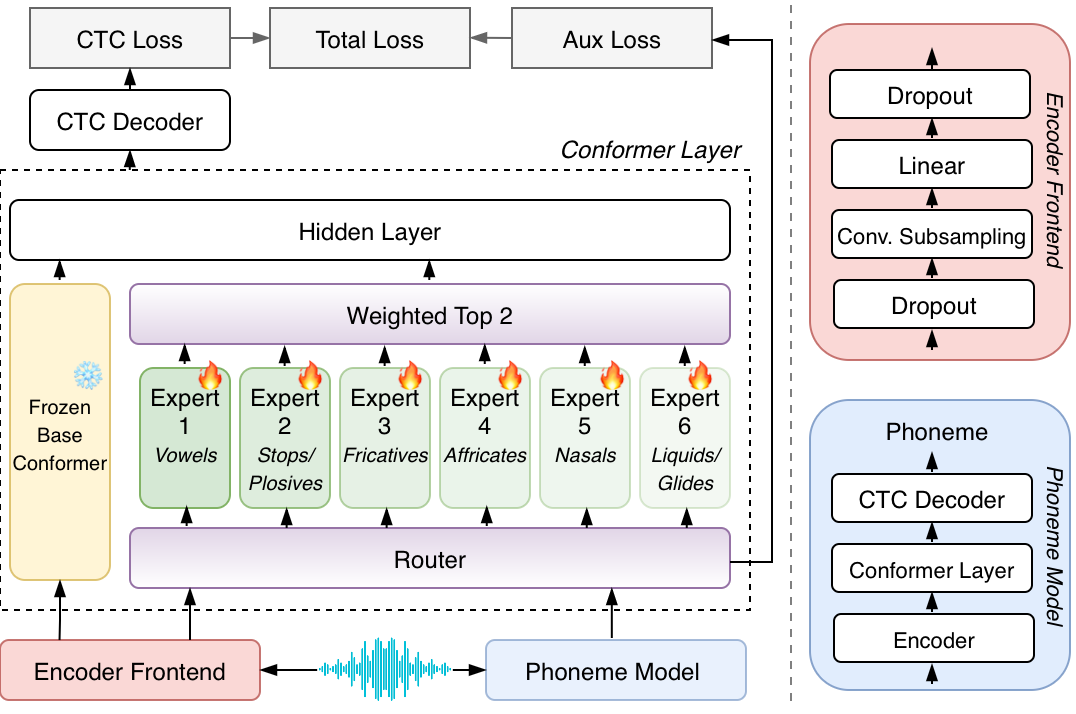}
  \vspace{-0.15 in}
  \caption{Architecture of MoPE-LoRA.}
  \label{fig:architecture}
  \vspace{-0.2 in}
\end{figure} 
%==================================================================================
\section{Proposed Method}
%==================================================================================

To handle phonetic variations in accented speech, we use MoPE-LoRA, which decomposes adaptation into linguistically motivated experts. Each expert targets a phonetic category (e.g., vowels, fricatives) via low-rank adapters, providing inductive bias for phonetic correction. Unlike conventional MoE, MoPE-LoRA leverages phonetic priors for focused adaptation (Figure~\ref{fig:architecture}).

% \subsection{Frozen Base Encoder}
% The backbone is a Conformer encoder~\cite{gulati2020conformerconvolutionaugmentedtransformerspeech} pretrained on large-scale speech corpora. Given an acoustic input $\mathbf{X} \in \mathbb{R}^{B \times T \times D}$, where $B$ is the batch size, $T$ is the number of frames, and $D$ is the feature dimension, the frozen encoder produces base representations $\mathbf{Y}_{\text{base}} = f_{\text{enc}}(\mathbf{X})$, where $f_{\text{enc}}$ denotes the full Conformer forward pass including convolution, multi-head self-attention, and feed-forward modules. These base representations preserve general acoustic knowledge while avoiding catastrophic forgetting of native speech patterns.

\subsection{Mixture of Phonetic Experts (MoPE) Architecture}

We group English consonants into five articulatory classes following standard manner-of-articulation distinctions~\cite{ladefoged2011course}, namely Stops/Plosives, Fricatives, Affricates, Nasals, and Liquids/Glides, and add Vowels as a sixth category, yielding six phonetic expert groups in total.
Table~\ref{tab:phoneme_to_expert} maps phonemes to the six linguistic expert categories used for routing and evaluation.

\begin{table}[h]
\centering
\caption{Phoneme-to-expert mapping for the 6 categories~\cite{ladefoged2011course}.}
\label{tab:phoneme_to_expert}
\vspace{-0.1 in}
\resizebox{0.8\columnwidth}{!}{
\begin{tabular}{cll}
\toprule
Idx & Category & Phonemes (ARPABET + stress) \\
\midrule
0 & Vowels & AA, AE, AH, AO, AW, AY, EH, ER,  \\
  &        & EY, IH, IY, OW, OY, UH, UW \\
1 & Stops & P, B, T, D, K, G \\
2 & Fricatives & F, V, TH, DH, S, Z, SH, ZH, HH \\
3 & Affricates & CH, JH \\
4 & Nasals & M, N, NG \\
5 & Liquids/Glides & L, R, W, Y \\
\bottomrule
\end{tabular}
}
\end{table}

This decomposition ensures that each expert is responsible for a coherent set of sounds that share articulatory properties, under the hypothesis that sounds produced by similar mechanisms are subject to related accent-induced distortions. Each group corresponds to a dedicated LoRA expert, allowing the model to learn specialized weight adjustments for different articulatory classes.

For a given encoder layer, the $e$-th expert introduces a low-rank weight perturbation to the frozen weight matrix $\mathbf{W}_{\text{base}}$:
\begin{equation}
\Delta W_e(\mathbf{x}) = \frac{\alpha}{r} \mathbf{B}_e \mathbf{A}_e \mathbf{x}, 
\quad e = 1, \dots, E,
\end{equation}
where $\mathbf{A}_e \in \mathbb{R}^{r \times D}$ is the down-projection matrix that maps the input from the original dimension $D$ into a low-rank subspace of dimension $r$, and $\mathbf{B}_e \in \mathbb{R}^{D \times r}$ is the up-projection matrix that maps back to $D$. The rank $r \ll D$ controls the capacity of each expert, and the scaling factor $\alpha$ modulates the magnitude of the adaptation relative to the frozen weights. $\mathbf{A}_e$ is initialized randomly and $\mathbf{B}_e$ is initialized to zero, so that at the start of training each expert contributes no perturbation and the model preserves its pretrained behavior.

Each of the $E=6$ experts specializes in an articulatory category, using separate low-rank subspaces to capture category-specific phonetic distortions. For example, the vowel expert models formant shifts and duration changes, while the fricative expert adapts to spectral variations in sibilants and non-sibilants across L1 backgrounds. This decomposition reduces interference seen in standard LoRA, where a single adapter must handle all phonetic categories within the same parameter space.

These expert modules are inserted into the Query ($\mathbf{W}_Q$), Key ($\mathbf{W}_K$), and Value ($\mathbf{W}_V$) projection matrices of the self-attention modules in adapted encoder layers. This placement is motivated by the fact that attention projections jointly govern how acoustic context is aggregated across frames, a process directly affected by phonetic distortion in accented speech. By adapting all three projections, the experts can modify both what the model attends to (via Q and K) and the content that is propagated (via V).

\subsection{Hybrid Phonetic Routing Mechanism}

To ensure that each expert specializes in its assigned phonetic category, we design a hybrid routing strategy that combines external linguistic supervision with internal acoustic learning.

A learnable gating network computes frame-level routing logits from the encoder's intermediate representations, producing learned routing probabilities:
\begin{equation}
\mathbf{p}^{\text{learn}}_{b,t} = \mathrm{softmax}\left(\frac{\mathbf{z}_{b,t}}{\tau}\right),
\end{equation}
where $\mathbf{z}_{b,t}$ are routing logits and $\tau$ is a temperature hyperparameter.

When phoneme labels $c_{b,t}$ are available from the frozen phoneme model (Section~\ref{sec:phoneme_model}), supervised routing provides a direct assignment signal:
\begin{equation}
\mathbf{p}^{\text{sup}}_{b,t,j} =
\begin{cases}
1, & c_{b,t} = j,\\
0, & c_{b,t} \neq j \text{ and } c_{b,t} \neq -1,\\
\frac{1}{E}, & c_{b,t} = -1,
\end{cases}
\end{equation}
where $j \in \{1, \dots, E\}$ indexes the experts, $c_{b,t} = j$ indicates that the frame belongs to phonetic category $j$, and $c_{b,t} = -1$ denotes low-confidence or silence frames that are routed uniformly. This supervised component acts as a hard constraint during training: frames confidently classified into phonetic category $e$ are deterministically assigned to expert $e$, ensuring that each expert is grounded in its intended articulatory class rather than learning arbitrary input partitions.

The final routing distribution combines both signals:
\begin{equation}
\mathbf{p}_{b,t} = \beta \mathbf{p}^{\text{sup}}_{b,t} + (1-\beta) \mathbf{p}^{\text{learn}}_{b,t},
\quad \beta \in [0,1],
\end{equation}
where $\beta$ controls the strength of phonetic supervision. The learned component captures acoustic patterns not fully explained by the phoneme model, such as coarticulation effects and phoneme boundary ambiguity characteristic of non-native speech~\cite{zsiga2003articulatory, zampini2008l2}, and can adaptively compensate for prediction errors on accented pronunciations, making the routing robust to the inherent imprecision of CTC-based alignment on accented speech.

Only the top-$k$ experts are activated per frame for computational efficiency:
\begin{equation}
\mathcal{K}_{b,t} = \mathrm{TopK}(\mathbf{p}_{b,t}, k).
\end{equation}
We set $k{=}2$ following ST-MoE~\cite{zoph2022stmoedesigningstabletransferable}, which demonstrated that top-2 routing achieves a good balance between expert specialization and model capacity. Additionally, $k{=}2$ allows frames near phoneme boundaries to activate experts from two relevant categories (e.g., a transition from a vowel to a fricative), which is particularly important for accented speech where phoneme boundaries may shift.

The selected expert outputs are combined via normalized routing weights:
\begin{equation}
\mathbf{Y}_{\text{moe}}[b,t] = \sum_{j=1}^{k} w_{b,t,j} \Delta W_{e_j}(\mathbf{h}_{b,t}),
\end{equation}
where $\mathbf{h}_{b,t}$ denotes the intermediate representation at frame $t$ of batch element $b$ produced by the current encoder layer, $e_j \in \mathcal{K}_{b,t}$ is the index of the $j$-th selected expert, and $w_{b,t,j}$ denotes its routing probability from $\mathbf{p}_{b,t}$ normalized over the top-$k$ activated experts. The final adapted representation is:
\begin{equation}
\mathbf{Y} = \mathbf{Y}_{\text{base}} + \mathbf{Y}_{\text{moe}}.
\end{equation}

Crucially, the expert set is shared across all accents. No accent-specific modules are introduced and no accent labels are required at test time; the model adapts to a given utterance purely through its frame-level phonetic routing decisions.

\subsection{Training Objectives}
The model is trained end-to-end using the Connectionist Temporal Classification (CTC) loss~\cite{graves2006ctc} as the primary objective:
\begin{equation}
\mathcal{L}_{\text{CTC}} = - \log P(\mathbf{y} \mid \mathbf{Y}).
\end{equation}

To stabilize MoE training, we incorporate two auxiliary losses. The first is a load balancing loss~\cite{fedus2022switchtransformersscalingtrillion}, which encourages uniform expert utilization by penalizing imbalanced routing:
\begin{equation}
\mathcal{L}_{\text{balance}} = E \sum_{i=1}^{E} f_i P_i,
\end{equation}
where $f_i$ and $P_i$ denote the assignment frequency and average routing probability of expert~$i$, respectively. The second is a router Z-loss~\cite{zoph2022stmoedesigningstabletransferable}, which discourages overconfident routing decisions by penalizing large routing probabilities:
\begin{equation}
\mathcal{L}_{\text{z}} = \frac{1}{BT} \sum_{b,t} \left( \max_i p_{b,t,i} \right)^2.
\end{equation}

The total training objective combines all three terms:
\begin{equation}
\mathcal{L}_{\text{total}} = \mathcal{L}_{\text{CTC}} + \lambda_{\text{balance}} \mathcal{L}_{\text{balance}} + \lambda_{\text{z}} \mathcal{L}_{\text{z}},
\end{equation}
where the auxiliary loss weights are set to $\lambda_{\text{balance}} = 0.01$ and $\lambda_{\text{z}} = 0.001$.

%==================================================================================
\section{Experimental Setup}
%==================================================================================

\subsection{Datasets}
We use the L2-ARCTIC dataset~\cite{Zhao2018L2ARCTIC}, containing approximately 24 hours of non-native English speech from 24 speakers (4 per accent) across Mandarin, Korean, Spanish, Hindi, Arabic, and Vietnamese L1 backgrounds. Following the evaluation protocol of~\cite{Mixture_LORA_Bagat_2025}, we enforce both speaker- and sentence-disjoint partitions across 4 splits, which is stricter than speaker-only disjointness as it prevents the model from memorizing sentence-level patterns during training. In each split, one speaker per accent (6 speakers) is held out for testing on sentences excluded from training, while the remaining 18 speakers are used for training and development. Speaker assignments rotate so that every speaker is evaluated exactly once, and results are averaged. Table~\ref{tab:dataset_stats_summary} reports per-accent statistics. For zero-shot evaluation, one accent is entirely excluded from training and used only for testing.

\begin{table}[H]
\vspace{-0.1 in}
  \caption{Dataset statistics by accent (averaged across 4 splits).}
  \label{tab:dataset_stats_summary}
  \centering
  \resizebox{0.62\columnwidth}{!}{
  \begin{tabular}{lccc}
    \toprule
    \textbf{Accent} & \textbf{Train} & \textbf{Val.} & \textbf{Test} \\
    \midrule
    Arabic       & 2h 37m & 19m & 6m \\
    Hindi        & 2h 18m & 17m & 5m \\
    Korean       & 2h 43m & 20m & 6m \\
    Mandarin     & 2h 51m & 21m & 7m \\
    Spanish      & 2h 48m & 21m & 7m \\
    Vietnamese   & 2h 52m & 21m & 7m \\
    \midrule
    \textbf{Total} & 16h 12m & 2h 03m & 41m  \\
    \bottomrule
  \end{tabular}
  }
  \vspace{-0.1 in}
\end{table}

For layer-wise representation analysis, we additionally use the TIMIT corpus~\cite{garofolo1993timit}, which provides time-aligned phonetic annotations.

\subsection{Training Configuration}
\label{sec:phoneme_model}
We use the NeMo Conformer CTC Small model (${\sim}$13M parameters) as the base encoder, with audio resampled to 16\,kHz. Models are trained for 50 epochs with batch size 16. MoPE-LoRA adapters with $E{=}6$ experts are applied to Q, K, V projections in layers 6--16, each with rank $r{=}8$ and $\alpha{=}16$. The routing network uses top-2 expert selection with hybrid routing (blend $\beta$ is a learnable parameter, initialized to $\sigma(0.5)$). The learning rate is $1 \times 10^{-3}$ with cosine annealing to $5 \times 10^{-6}$. Training is conducted on an NVIDIA RTX 5090.

\textbf{Phoneme Model}.
To provide frame-level phoneme supervision for expert routing, we train a separate phoneme-level CTC model on LibriSpeech~\cite{panayotov2015librispeech}, using a Conformer CTC Small architecture (16 layers, dimension 176). Grapheme-to-phoneme conversion uses \texttt{g2p-en}. The model is trained for 25 epochs with Adam ($\beta_1{=}0.9$, $\beta_2{=}0.999$, weight decay $1 \times 10^{-5}$), learning rate $1 \times 10^{-3}$ with cosine annealing and 10\% warmup. During MoPE-LoRA training, this model remains frozen.

% We use 80-dimensional log-mel spectrograms. The output vocabulary consists of 70 ARPABET phonemes generated using \texttt{g2p-en}, including lexical stress markers, along with a CTC blank symbol. Training is performed on the full LibriSpeech 960-hour set, without SpecAugment. Optimization uses Adam ($\beta_1{=}0.9$, $\beta_2{=}0.999$, weight decay $1 \times 10^{-5}$) with an initial learning rate of $1 \times 10^{-3}$, cosine annealing scheduling, and a batch size of 16. The best checkpoint is selected based on validation loss. For inference, we use greedy CTC decoding.

\section{Results and Analysis}
\subsection{Phoneme Model Evaluation}
\begin{table}[H]
\centering
\vspace{-0.2 in}
\caption{Phoneme Error Rate (PER) on LibriSpeech subsets.}
\vspace{-0.1 in}
\label{tab:per_librispeech}
\resizebox{0.38\columnwidth}{!}{
\begin{tabular}{lcc}
\toprule
\textbf{Subset} & \textbf{PER (\%)} \\
\midrule
dev-clean  & 2.81 \\
test-clean & 2.92 \\
dev-other  & 8.02 \\
test-other & 7.93 \\
\bottomrule
\end{tabular}}
\vspace{-0.1 in}
\end{table}
We first evaluate our phoneme model on LibriSpeech to establish a native-English baseline. As shown in Table~\ref{tab:per_librispeech}, the results confirm strong performance on both clean and noisy speech. 

\begin{table}[h]
\centering
\vspace{-0.1 in}
\caption{Frame-level phonetic category accuracy of the frozen phoneme model on L2-ARCTIC cross-domain dataset.}
\vspace{-0.1 in}
\resizebox{0.62\columnwidth}{!}{
\label{tab:phoneme_accuracy}
\begin{tabular}{lcc}
\toprule
\textbf{Category} & \textbf{Top-1 (\%)} & \textbf{Top-2 (\%)} \\
\midrule
Vowels          & 32.70 & 47.83 \\
Stops/Plosives  & 39.18 & 75.74 \\
Fricatives      & 35.36 & 72.35 \\
Affricates      & 25.35 & 45.80 \\
Nasals          & 42.33 & 67.34 \\
Liquids/Glides  & 41.57 & 70.06 \\
\midrule
\textbf{Overall} & \textbf{36.04} & \textbf{61.56} \\
\bottomrule
\end{tabular}}
\vspace{-0.15 in}
\end{table}

Cross-domain evaluation on L2-ARCTIC reveals a substantial drop in frame-level phonetic accuracy, with 36.04\% top-1 and 61.56\% top-2 accuracy (Table~\ref{tab:phoneme_accuracy}), highlighting the challenge of cross-domain accented phoneme recognition~\cite{mimura25_interspeech}. This motivates our $k{=}2$ expert routing strategy: dual expert activations help capture coarticulatory transitions and provide smoother gradient flow than strict top-1 sparse routing~\cite{muqeeth24_smear}. Combined with the learned gating ($1{-}\beta$), this approach achieves 10.43\% WER on accented speech (Table~\ref{tab:mope_results}), demonstrating effective compensation for cross-domain phonetic variability.

%==================================================================================

%==================================================================================

\subsection{Layer Selection Strategy}

Not all encoder layers contribute equally to phonetic adaptation. Prior studies suggest that early layers capture low-level acoustic features, middle layers encode most phonetic information, and upper layers focus on contextualized representations~\cite{pasad2022layerwiseanalysisselfsupervisedspeech}.

\begin{figure}[H]
    \centering
    \includegraphics[width=0.72\linewidth]{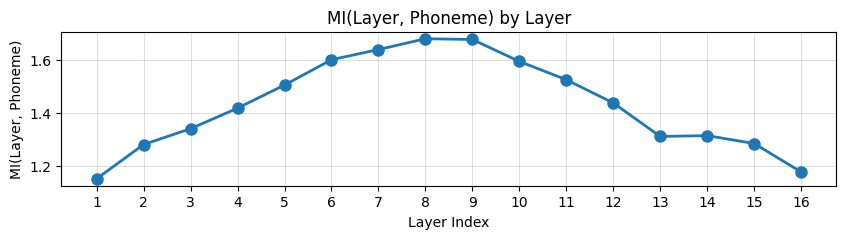}
    \vspace{-0.1 in}
    \caption{Mutual information between layer representation and phoneme label.}
    \label{fig:phonetic_analysis}
    \vspace{-0.1 in}
\end{figure}

\begin{figure*}[t]
\centering
\includegraphics[width=1\linewidth]{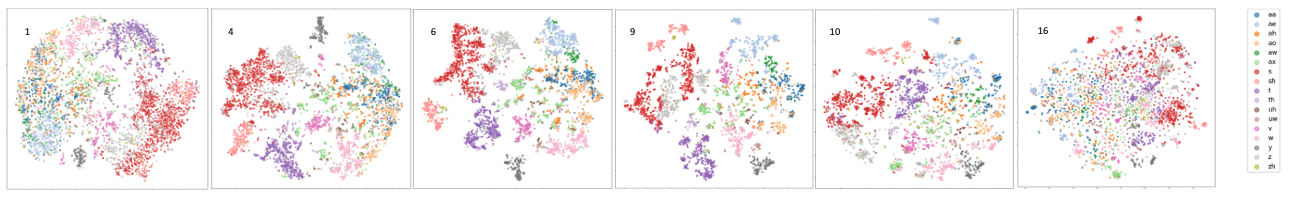}
\vspace{-0.25 in}
\caption{t-SNE visualizations of phoneme embeddings from representative encoder layers (1, 4, 6, 9, 10, 16). Each point is a frame colored by phoneme label. Cluster structure is clearest in the middle layers and more diffuse at Layer 16, consistent with the MI trend in Figure~\ref{fig:phonetic_analysis} and supporting our choice to focus adaptation on layers 6--16.}
\label{fig:phoneme_tsne}
\vspace{-0.15 in}
\end{figure*}

To verify this, we computed the mutual information (MI) between each Conformer layer and phoneme labels on TIMIT. Continuous embeddings were discretized via K-Means, and MI was calculated between cluster assignments and ground-truth phonemes (Figure~\ref{fig:phonetic_analysis}). The MI shows an inverted-U trend: middle layers contain the most phonetic information, while early and upper layers contain less. t-SNE visualizations (Figure~\ref{fig:phoneme_tsne}) confirm this pattern.

Based on these results, we focus phonetic adaptation on layer ranges that include the middle, phoneme-rich layers, while varying the inclusion of early and upper layers.

\subsection{Multi-Accent ASR}

Table~\ref{tab:mope_results} reports WER across adaptation strategies and layer ranges. Among Single LoRA variants, adapting all three attention projections (QKV) across layers 1--16 achieves the best WER (11.33\%), outperforming partial-projection LoRA and Full Fine-Tuning. Limiting adaptation to upper layers degrades performance, highlighting the importance of middle layers for phonetic correction.

MoPE-LoRA consistently surpasses Single LoRA across all tested layer ranges. The best configuration, MoPE-QKV (6--16), achieves 10.43\% WER—a 2.37\% absolute improvement over Full FT and 0.90\% over Single LoRA-QKV (1--16)—while using only 13.6M parameters. This demonstrates that phoneme-category experts provide a linguistically motivated decomposition that generalizes across accents more efficiently than the 244M-parameter accent-level MAS-LoRA~\cite{Mixture_LORA_Bagat_2025}.

Consistent with our MI (Figure~\ref{fig:phonetic_analysis}) and t-SNE (Figure~\ref{fig:phoneme_tsne}) analysis, Layers 1--5 encode low-level, accent-invariant acoustic features contributing little to phonetic adaptation. Performance improves steadily as middle layers (6--12) are included, reflecting the emergence of phoneme-discriminative representations. Despite lower phonetic MI, late layers (13--16) remain essential for the contextual representations required by CTC decoding; extending adaptation to all layers (1--16) offers no gain over 6--16, and restricting to middle layers alone (4--12) proves detrimental.

\begin{table}[t]
\centering
\caption{WER (\%) on L2-ARCTIC with different adaptation methods. Total Params includes all experts and routing; Used Params denotes parameters active during inference.}
\vspace{-0.1 in}
\label{tab:mope_results}
\resizebox{\columnwidth}{!}{
\begin{tabular}{lcccc}
\toprule
\textbf{Method} & \textbf{Layers} 
& \shortstack{\textbf{Total} \\ \textbf{Params (M)}} 
& \shortstack{\textbf{Used} \\ \textbf{Params (M)}} 
& \textbf{WER (\%)} \\
\midrule
No FT              & --     & 13.0 & 13.0 & 30.82    \\
Full FT            & all    & 13.0 & 13.0 & 12.80   \\
MAS-LoRA~\cite{Mixture_LORA_Bagat_2025} & -- & $\sim 244.0$ & $\sim 9.2$ & 11.77 \\
\midrule
Single LoRA-QK     & 1--16  & 13.1 & 13.1 & 14.47    \\
Single LoRA-QV     & 1--16  & 13.1 & 13.1 & 11.72    \\
Single LoRA-KV     & 1--16  & 13.1 & 13.1 & 12.46    \\
Single LoRA-QKV    & 1--16  & 13.1 & 13.1 & 11.33    \\
Single LoRA-QKV    & 7--16  & 13.1 & 13.1 & 11.55    \\
Single LoRA-QKV    & 8--16  & 13.1 & 13.1 & 12.15    \\
\midrule
MoPE-QKV           & 8--16  & 13.5 & 13.2 & 11.03   \\
MoPE-QKV           & 7--16  & 13.5 & 13.2 & 10.85   \\
MoPE-QKV           & 6--16  & 13.6 & 13.2 & \textbf{10.43}  \\
MoPE-QKV           & 1--16  & 13.8 & 13.3 & \textbf{10.43}   \\
MoPE-QKV           & 4--12  & 13.6 & 13.2 & 14.05   \\
MoPE-QKV           & 4--16  & 13.7 & 13.3 & 10.67   \\
\bottomrule
\end{tabular}}
\vspace{-0.15 in}
\end{table}

\subsection{Zero-Shot Generalization to Unseen Accents}
\begin{table}[h]
\vspace{-0.1 in}
\centering
\setlength{\tabcolsep}{4pt}
\caption{Zero-shot WER (\%) on accents unseen during training. AR: Arabic, ZH: Mandarin, HI: Hindi, KR: Korean, SP: Spanish, VI: Vietnamese. Bold: best per accent.}
\vspace{-0.1 in}
\label{tab:zero_shot_accent}
\resizebox{0.78\columnwidth}{!}{
\begin{tabular}{lccccccc}
\toprule
\textbf{Method}  & AR & ZH & HI & KR & SP & VI & Mean \\
\midrule
No FT & 17.98 & 20.56 & 9.28 & 13.47 & 17.08 & 29.24 & 17.93 \\
Full FT  & 12.48 & 14.25 & 7.21 & 7.90 & 11.88 & 21.79 & 12.58 \\
LoRA  & 10.87 & 13.05 & 7.34 & 6.66 & 10.41 & 19.97 & 11.38 \\
MoPE  & \textbf{9.56} & \textbf{11.76} & \textbf{6.01} & \textbf{5.21} & \textbf{9.21} & \textbf{18.14} & \textbf{9.98} \\
\bottomrule
\end{tabular}}
\vspace{-0.1 in}
\end{table}
As shown in Table~\ref{tab:zero_shot_accent}, MoPE achieves the lowest WER on all unseen accents, with a 12.3\% relative improvement over Single LoRA and 44.3\% over the unadapted baseline (per-accent reductions on range: 9.2\%–21.8\%). Without accent-specific experts or labels, phoneme-category specialization addresses pronunciation variability at its root, enabling robust zero-shot generalization.

\section{Conclusion}

We introduce MoPE-LoRA, a Mixture of Phonetic Experts framework adapting Conformer ASR to accented English via phoneme-category experts and hybrid frame-level routing, with no accent-specific modules required. On L2-ARCTIC, it achieves 10.43\% WER and 44.3\% relative improvement over the unadapted baseline in zero-shot transfer, confirming phoneme-level decomposition as a principled, parameter-efficient solution for accent-agnostic ASR.

\section{Acknowledgments}
This work was supported by the BRAINS project
(PID2024-155948OB-C52) funded by
MCIN/AEI/10.13039/501100011033 and, as appropriate,
by ``ERDF A way of making Europe'' and by the
``European Union,'' and by the Program
`ACTIVIDADES I+D PROCESOS HUMANOS Y SOCIALES'
of the Comunidad de Madrid, Spain, under Project
PHS2024/PH-HUM-52 (Innovatrad-CM).

\section{Generative AI Use Disclosure}
Generative AI tools were used for proofreading and
improving the clarity of the manuscript. All research
ideas, experimental design, implementation, and
analysis were conducted entirely by the authors.

\bibliographystyle{IEEEtran}
\bibliography{mybib}

\end{document}